\documentclass[letterpaper]{article}
\usepackage{aaai}
\usepackage{times}
\usepackage{helvet}
\usepackage{courier}
\usepackage{graphicx}
\usepackage[colorinlistoftodos]{todonotes}

\begin{document}
%
\title{A Train Status Assistant for Indian Railways}
\author{Himadri Mishra{\tiny\dag} and Ramashish Gaurav{\tiny\ddag} and Biplav Srivastava{\tiny\P}\\
{\tiny\dag}Whodat, India; {\tiny\ddag}Nutanix, India;  {\tiny\P}IBM, USA
}
\maketitle
\begin{abstract}
\begin{quote}
Trains are part-and-parcel of every day lives in countries with large, diverse, multi-lingual population like India. Consequently, an assistant which can accurately predict and explain train delays will help people and businesses alike. We present a novel conversation agent which can engage with people about train status and inform them about its delay at in-line stations. It is trained on past delay data from a subset of trains and generalizes to others.
\end{quote}
\end{abstract}

\section{Introduction}

Trains provide an inexpensive form of inter-city and long-distance movement for people and goods in many countries around the world like India. To keep them well maintained and provide good service to travelers, AI is seen as a promising new technology \cite{train-ai}. 
However, in developing countries, trains often suffer from endemic delays which can be credited to obsolete technology, large network size and external factors like weather. 

Consider train numbers 12305 and 13050 on Indian Railways\footnote{https://etrain.info/in?TRAIN=12305, https://etrain.info/in?TRAIN=13050. Trains are often late by tens of hours or even days in extreme weather.}. Train 12305, Kolkata Rajdhani, is a premium train which runs for nearly 18 hours to cover 1529 km (in 5 states that speak Hindi, Bengali and English), stops at 
10 stations and was late over 1 hour at destination for 19\% of the days  between 1-Oct-2017 and 1-Sep-2018. Train 13050, Amritsar Howrah Express, runs for nearly 46 hours over 3 days to cover 1922 km (in 7 states that speak Hindi, Bengali, Punjabi and English), stops at 112 stations and was late over 1 hour at destination for 63\% of the days  during the same period. 

In this context, travelers often want to know when a train may arrive at a station, whether it is delayed and what alternative travel options (like another lesser delayed train) they may have and in their own languages. Currently, people are given delay information about trains from over-crowded public booths by human agents or search-interfaces on websites 
in English (which is spoken by a small segment of train-riding population), systems are not aware of a user's context and they 
provide  delay estimates incrementally which usually increases as time elapses; making re-planning hard.

There has been some recent efforts to understand these delays and predict them after learning from past delay data  in India \cite{train-estimation}, Serbia \cite{train-estimation-serbia} and USA \cite{train-estimation-us}. But can these analytic model help travelers and other stakeholders (e.g., businesses serving passengers, people affected by travelers, train managers) who need guidance for their activities? 


In response, we propose a conversation interface for common man to inquire about train delay which may further be personalized to user's need \cite{personal-chatbot}. This gives some unique benefits:
(1) {\bf New Temporal insights}: Currently, methods have delay information of on-going journeys but no estimate on the probable delays for journeys in upcoming weeks or months, or past performance. This does not enable travelers to schedule their journey plans efficiently with sufficient time in hand and hence, cause frustration.
(2) {\bf New Journey insights}: 
Apart from delays at a certain station, people want to know (a) if a train will be delayed further in journey, (b)  the bottleneck station which becomes the root cause of the delay, 
and (c) if the delay gets mitigated or not during travel, 
(d) the delay characteristics of trains and routes, (e) similarity of trains based on delays they encounter. 
(3) {\bf New Interaction Mode}: The conversation interface allows people to access information without knowledge of computer systems which can be personalized to their need and language, and is aware of their traveling context. 



\section{Example Illustration}

A sample example is shown in Figure~\ref{fig:example}. The user starts by asking about a train without specifying the location and date for which she is interested in. The agent assumes the intended location is 
destination station on its journey and date as current, and responds in natural language with the predicted delay. It also inquires if the user's request is satisfied. The user specifies another location which the agent now satisfies.  The user wants to know the first location for delay and the system checks the stations on the journey to know first occurrence of delay. 


\begin{figure}
\center{
\fbox{
\begin{minipage}[t]{4in}
{\tiny 
\begin{tabbing} 
{\bf User:}\=          // Turn 1 \\
   \> Is train 12307 on time? \\
{\bf Agent:}\= // Answering assuming destination \\
   \> Train Number 12307 will be delayed by 159 minutes at  \\
   \> JU station  on 2018-09-21. \\
{\bf User:}\=          // Turn 2 \\
   \> How about for Varanasi? \\  
{\bf Agent:}\=  // Handling error condition \\
   \> Train 12307 does not stop at Varanasi. \\  
   \> Here is the list of stations on this route. ...\\
{\bf User:}\=          // Turn 3 \\
   \> No, I meant for Allahabad. \\
{\bf Agent:}\= // User-specified station \\
   \> Train 12307 will be delayed further after station ALD\\ 
   \> on 2018-09-21 by 110.8 minutes \\
{\bf User:}\=          // Turn 4 \\
   \> What is the average train delay? \\  
{\bf Agent:}\=  ... // Route analytics  \\
\end{tabbing}
}
\end{minipage}
}}
\caption{\label{fig:example} A sample conversation.; video at https://youtu.be/a-ABv29H6XU}
\vspace{-10pt}
\end{figure}

\section{System Details}


\begin{figure}
 \centering
   \includegraphics[width=0.35\textwidth]{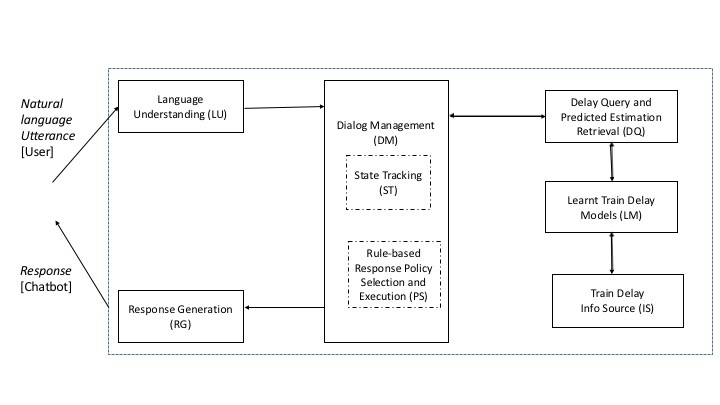}
  \caption{The architecture of train assistant.}
  \label{fig:chatbot-arch}
\vspace{-20pt}
\end{figure}

\noindent {\bf System Architecture:} Figure~\ref{fig:chatbot-arch} shows the working of system  at conceptual level. 
Here, the user's utterance is analyzed to detect her intent and a 
policy for response is selected. This policy may call for predicting a given train's delay at a given station and time, and result is returned which is used by response generator to create a response using templates. The system can select defaults for missing values, get prediction from previously trained models,  and decide if it has a reasonable estimate (timely value and enough confidence in an estimate's correctness) to respond.

In principle, the chatbot could have been implemented in any framework since the call to predict delays is a REST call to cloud-based service hosting the learned predictive models. However, in practice, the duration of backend call  varies a lot depending on the nature of query and train, and the behavior of different chatbot platforms is not the same. We experimented with Google's DialogFlow and Microsoft's Bot Framework, and eventually chose the latter. 

\noindent {\bf Learned Models:} The details of data collection and model training are given in \cite{train-estimation}. We collected  delay information of 135 Indian trains over two years from March 2016 to February 2018, and categorized the 135 trains into two groups based on the amount of data collected: 52 \textit{Known Trains} (52KT) for large and 83 \textit{Unknown Trains} (83UT) for small sets. We trained Random Forest Regressor (RFR) and Ridge Regressor (RR) models for each station en-route in 52KT's journey \textit{Known Station}, and learned a late minutes prediction framework which could 
generalize to 83UT too.
We conducted exhaustive experiments in multiple settings and chose Confidence Intervals (CI) to articulate aggregated accuracy results. Adopting the popular CI classes of 68\%, 95\% and 99\%, we achieved 28\%, 56\% and 67\% accuracy in the respective classes for 52KT's test data. The trained model used in the demo corresponds to RFR at 99\% CI which correspond to highest accuracy obtained.


\begin{figure}
 \centering
   \includegraphics[width=0.3\textwidth]{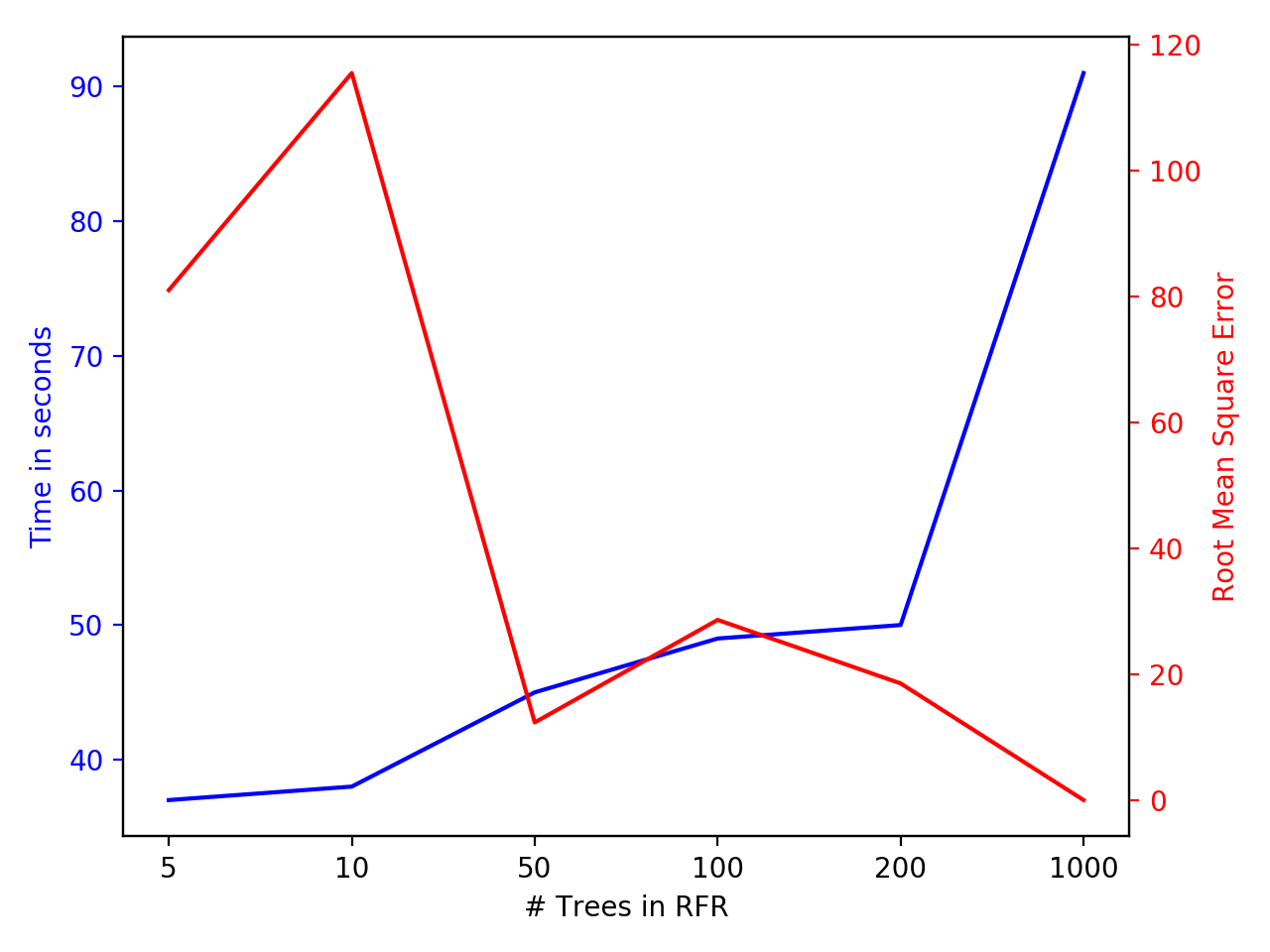}
  \caption{The time - accuracy (in RMSE) trade-off for train 13050 with largest number of stations in its journey}
  \label{fig:time_acc_toff}
\vspace{-10pt}
\end{figure}

Since the learned models are used by a chatbot which needs to converse in real-time, apart from accuracy, we also consider the response time of models to predict delays for new trips.  The performance of RR depends only on the number of stations of the train and took 20-22 seconds to predict  late minutes for the longest train we considered - Train 13050 with 112 stations.
RFR was more accurate; for it, 
we experimented with the number of stations a train has, the number of trees in the RFR, the time to predict and  Root Mean Square Error (RMSE). In Figure~\ref{fig:time_acc_toff}, we show the result for Train 13050. The prediction time increases linearly with size of forest while error reduces sharply. 
We find that 5-50 trees RFR give reasonable trade-off  and use it in the system. 
\vspace{-12pt}
\section{Discussion}






We demonstrated a novel conversation agent using AI methods to help common man gain insights about train delays and work around them. We used a slot-based dialog approach for front-end and a learning-based predictor trained on 2 years of training data. The system maintains user context and supports analytical queries.
The system can be easily extended by enhancing the scope of train data (number, history), the learning models and languages for conversation. 

\vspace{-10pt}


\bibliographystyle{aaai}
\bibliography{main}

\begin{thebibliography}{}

\bibitem[\protect\citeauthoryear{Daniel \bgroup et al\mbox.\egroup
  }{2018}]{personal-chatbot}
Daniel, F.; Matera, M.; Zaccaria, V.; and Dell'Orto, A.
\newblock 2018.
\newblock Toward truly personal chatbots: On the development of custom
  conversational assistants.
\newblock In {\em Proc. 1st Intl Wk. Software Engg. for Cognitive Services},
  SE4COG '18,  31--36.

\bibitem[\protect\citeauthoryear{Gaurav and
  Srivastava}{2018}]{train-estimation}
Gaurav, R., and Srivastava, B.
\newblock 2018.
\newblock Estimating train delays in a large rail network using a zero shot
  markov model.
\newblock {\em IEEE Intl. Conf. Intelligent Transportation (ITSC), Hawaii}.

\bibitem[\protect\citeauthoryear{Markovic \bgroup et al\mbox.\egroup
  }{2015}]{train-estimation-serbia}
Markovic, N.; Milinkovic, S.; S.~Tikhonov, K.; and Schonfeld, P.
\newblock 2015.
\newblock Analyzing passenger train arrival delays with support vector
  regression.
\newblock In {\em TRB Part C}, volume~56.

\bibitem[\protect\citeauthoryear{Rossi}{2015}]{train-ai}
Rossi, B.
\newblock 2015.
\newblock Trains with brains: how artificial intelligence is transforming the
  railway industry.
\newblock {\em Information Age,
  www.information-age.com/trains-brains-how-artificial-intelligence-transforming-railway-industry-123460379/}.

\bibitem[\protect\citeauthoryear{Wang and Work}{2015}]{train-estimation-us}
Wang, R., and Work, D.~B.
\newblock 2015.
\newblock Data driven approaches for passenger train delay estimation.
\newblock In {\em 2015 IEEE 18th Intl. Conf. on Intelligent Transportation
  Systems},  535--540.

\end{thebibliography}

\end{document}